\documentclass[11pt,a4paper]{article}
\usepackage[hyperref]{acl2017}
\usepackage{times}
\usepackage{latexsym}
\usepackage{amsmath}
\usepackage{amssymb}
\usepackage{algorithm}
\usepackage{algorithmic}
\usepackage{multirow}
\usepackage{graphicx}
\usepackage{url}

\usepackage{times}
\usepackage{epsfig}
\usepackage{graphicx}
\usepackage{amsmath}
\usepackage{amssymb}
\usepackage{times}
\usepackage{epsfig}
\usepackage{graphicx}
\usepackage{amsmath}
\usepackage{amssymb}
\usepackage{times}
\usepackage{helvet}
\usepackage{courier}
\usepackage{amsmath,amssymb}
\usepackage{algorithm}
\usepackage{algorithmic}
\usepackage{multirow}
\usepackage{graphicx}
\usepackage{color}
\usepackage{marvosym}
\usepackage{makecell}

\usepackage{url}

\aclfinalcopy 

\title{Feature extraction and evaluation for BioMedical Question Answering}

\author{ Ankit Shah, Srishti Singh, Shih-Yen Tao \\
  {\tt {aps1,srishti1,shihyent}@andrew.cmu.edu} \\ 
 \textbf{Course Instructor - Prof.Eric Nyberg}}

\date{}

\begin{document}
\maketitle

\section*{Abstract}
In this paper, we present our work on the BioASQ pipeline. The goal is to answer four types of questions: summary, yes/no, factoids, and list. Our goal is to empirically evaluate different modules involved: the feature extractor and the  sentence selection block. We used our pipeline to test the effectiveness of each module for all kind of question types and perform error analysis. We defined metrics which are useful for future research related to the BioASQ pipeline critical to improve the performance of the training pipeline. 

\section{Introduction}
The biomedical semantic question answering task - task 6b  comprises a large scale question answering challenge, which is given for a systems answer at all stages of question answering task. There are four types of bio medical questions - yes/no, factoid, list and summary questions. The system under test is provided with English questions written by biomedical experts which reflect real-life information requirements.

\begin{figure*}[h]
    \centering
    \includegraphics[width=\textwidth]{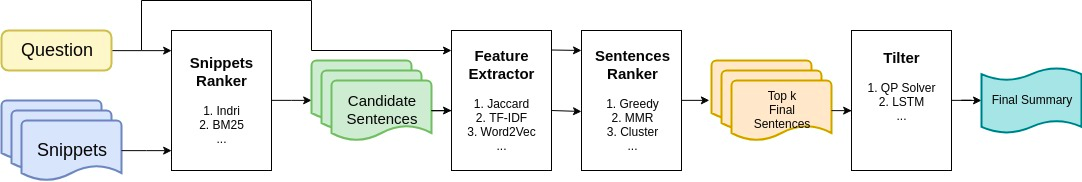}
    \label{fig:pipeline}
    \caption{The Overall Pipeline based on important modules needed for question answering}
\end{figure*}

The overall pipeline is described in Fig~\ref{fig:pipeline}. For each question, a list of supporting documentations are also given. At first, a snippets ranker is used to select snippets which are relevant to the question. Secondly, a sentence ranker along with the feature extractor module perform ranking algorithm to choose top $k$ candidate sentences for the final summary. In the end, the tiler module gather all sentences and compress them into the final result.

To be more specific, the snippets ranker use the Indri algorithm~\cite{W18-2312} to choose relevant snippets, and the tiler module used quadratic programming to predict the final output. As for the feature extractor module, which we test in this project, use either Jaccard, Dice, Tf cosine similarity~\cite{featureextractors}, dependency parser~\cite{chen2014} or the combined feature. For the sentence selection module, we used a greedy approach as the baseline and compared it with the MMR algorithm~\cite{carbonell1998use}.

\section{Dataset Description}
\begin{figure}
    \centering
    \includegraphics[width=0.5\textwidth]{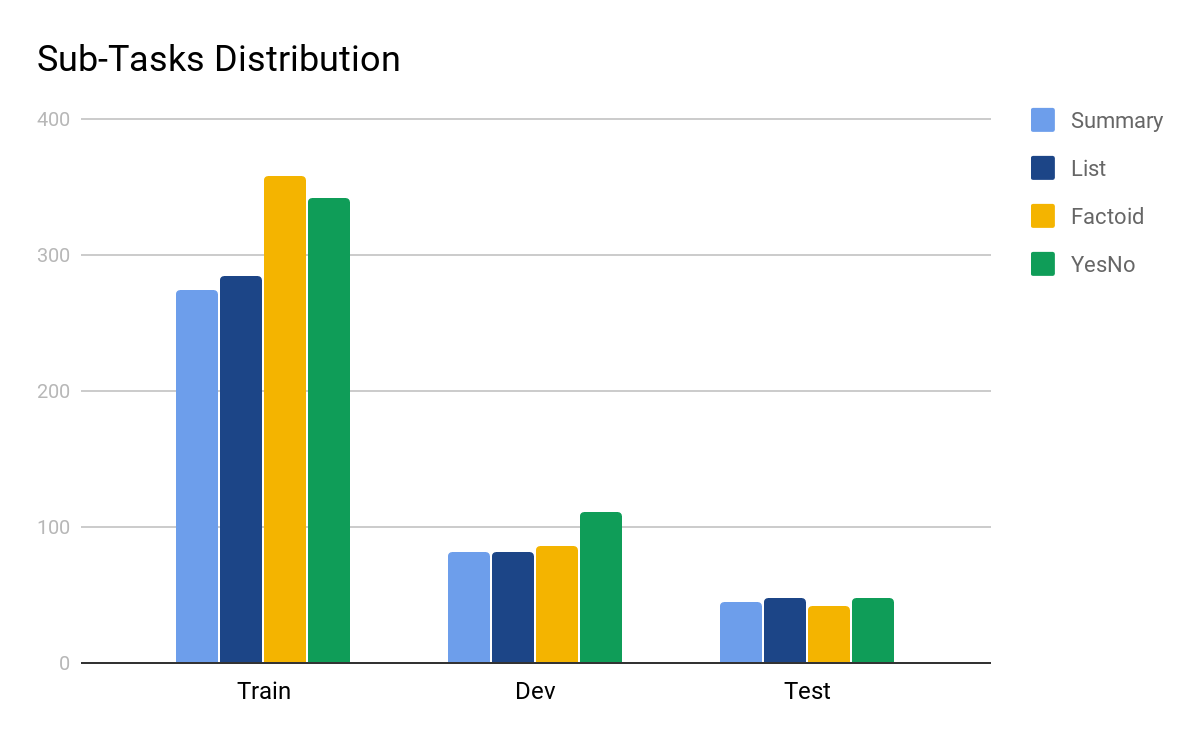}
    \caption{Distribution of Questions in the Dataset.}
    \label{fig:distribution}
\end{figure}

In this section we introduce the details of the BioASQ dataset. In order to evaluate our pipeline, we divided the training data provided by the BioASQ challenge into training/dev/test set with the ratio of $7:2:1$. The whole dataset consists of four sub-types of questions: list, factoid,yes/no, and summary. The first three come along with one ideal answer and one exact answer, while the summary question only have an ideal answer. Note that in our project, we evaluate the ideal answer for the summary question and the exact answer for the other three types of questions. The overall statistics and the distribution of the questions are summarized in Table~\ref{tbl:dataset} and Fig~\ref{fig:distribution}, respectively.

\begin{table}[]
\centering
\caption{Dataset Statistics.}
 \resizebox{0.98\columnwidth}{!}{
\begin{tabular}{|l|l|l|l|}
\hline
\textbf{Question Types}        & \textbf{Training} & \textbf{Dev} & \textbf{Test} \\ \hline
Total   & 1259     & 359 & 181  \\ 
Summary & 274      & 81  & 45   \\ 
List    & 285      & 81  & 47   \\ 
Factoid & 358      & 86  & 42   \\
Yes/No  & 342      & 111 & 47   \\ \hline
\end{tabular}}
\label{tbl:dataset}
\end{table}

\section{Similarity evaluation metrics}

Jaccard Similarity: - We  calculate the Jaccard similarity, which is the intersection over union and it is a statistic used for comparison of the similarity and diversity of the sample sets. \\

TF-IDF with word embeddings : - Ontologies like WordNet and UMLS/SNOMEDCT are used in biomedical domain for concept expansion for incorporating the semantics while computing the sentence similarity is not sufficient due to the nature of the ontologies. Thus, semantic information is more controlled and updated with a similarity metric inspired by TF-IDF~\cite{chandu2017tackling}. This similarity metric uses the symmetric word to word similarity matrix and the tf-idf vectors for the sentences among which the similarity is computed.  \\

Dice coefficient : - Dice coefficient is similar to the Jaccard index however one takes the string measure as - say there are two strings x and y then using the bigrams the string similarity is computed as s = 2*$n_t$/$n_x$ + $n_y$ \\

Cosine correlation similarity :- Cosine similarity has an interpretation of the angle between the two vectors. 

\section{Approach}
We aim to address the research questions on comparing different feature extractor and sentence ranking algorithm for the Biomedical question answering system. We run our pipeline on both ideal and exact question evaluation. The details of each component is introduced in the following.

The feature extractor can be interpreted as a similarity function in fact. Given two sentences $A,B$, the module $f(A,B)$ output a similarity score of the pair. The Jaccard feature extractor~\cite{Kobayakawa:2009:FCS:1697519.1697624} computes the score by $f(A,B) = \frac{|A \cap B|}{| A \cup B|}$, which computes the overlapping ratio of two sentences. The Dice feature extractor~\cite{W16-4118} does a similar thing by computing $f(A,B) = \frac{2 | A \cup B|}{|A|+|B|}$. On the other hand, the TF module first convert each sentence into a TF feature representation $tf(A), tf(B)$, and computes the cosine similarity between them. The above three module ends up with three scores, the combined feature extractor simply outputs the score sum.

\begin{figure}
    \centering
    \includegraphics[width=0.5\textwidth]{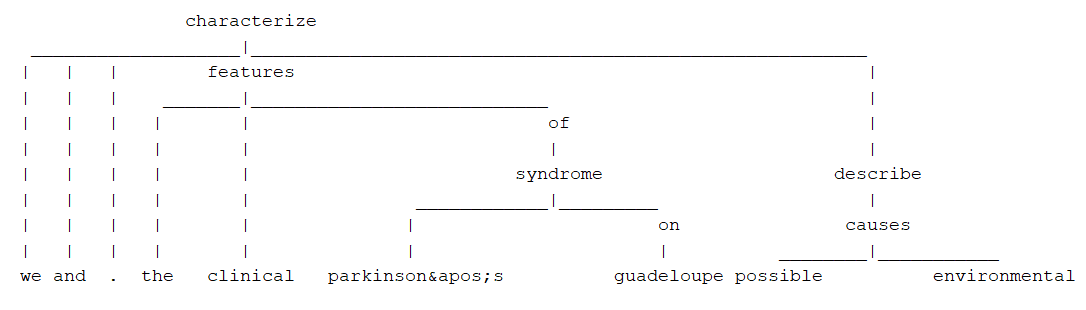}
    \caption{Dependency Parsed Sentence Structure}
    \label{fig:dependency}
\end{figure}

A dependency parser analyzes the grammatical structure of a sentence, establishing relationships between "head" words and words which modify those heads. Such relations are important to establish to understand the extent to which a head word is affected by words surrounding them. The use of such a dependency parser is missing in BioMedical question answering literature. Since uncovering the semantic structure and extent of dependencies is critical to answer questions when humans perform question answering, we think it will improve and play a critical role in the BioMedical Question answering pipeline, thus we experiment using the dependency parser. We implemented the dependency parser as a feature in combination with Jaccard, cosine and tf-idf similarity.
The parser works with sentence roots to draw similarity between question and candidate sentences Fig~\ref{fig:dependency}.

For the sentence selection module, the baseline we are going to try is the greedy selections. That is to say, given we can at most select $k$ sentence for final summary, we greedily select top-k sentences with highest score. Or course, this naive strategy results in poor summary result, since it introduces lots of redundancy, which is certainly not a desirable property for high quality summary. As the result, our main goal for this particular module is to reduce the sentence redundancy.

As the result, we leveraged Maximal Marginal Relevance ~\cite{carbonell1998use} as the additional approach. In a nutshell, a candidate is selected based on the similarity score with {\emph BOTH} question sentence and another candidate sentences which were already selected as answer. As shown in previous literature, MMR could effectively reduce the redundancy.

\section{Error analysis and metrics}
We use the standard ROUGE scores~\cite{rouge} for evaluating the pipeline. Recall-Oriented Understudy for Gisting Evaluation (ROUGE) has various types of performance analysis metrics. For our task we use ROUGE-l as the metric for evaluation. ROUGE-l is the longest common subsequence based statistics where the longest common subsequence takes into account sentence level structures similarity naturally and identifies the co-occuring sequence n-grams automatically. Such a metric will enable us to answer the questions and thus it is useful for the BioASQ question answering summary evaluation task where the predicted answer and the ideal answer will be used as references to evaluate the ROUGE metric.
In order to understand the accuracy of the system and post process the output, we defined two new metrics to evaluate the end to end system. We know that there are questions such as yesno question type, list question type and factoid question types whose ideal answer as well as exact answers are defined in the system. We want our system to not only perform well on not only the summary type questions but also all other types of questions. This is mainly a post processing step which will enable further understanding of the system and can be applied for the cases where the questions with exact answer information is already available. These metrics are as follows. 
\begin{itemize}
    \item Soft exact answer based accuracy measure - This measure is defined to calculate accuracy of the system when any of the exact answers is found in the predicted answer in the overall system. 
    \item Hard exact answer based accuracy measure - This measure is defined to calculate accuracy of the system when all of the exact answers is found and predicted in the overall system. 
\end{itemize}

\section{Experimental Design}
Our design would be based on the OAQA system extension as well as using the BioASQ - Rabbit pipeline in order to use the overall pipeline and extend the same by adding functionality for each module namely, Feature Extractor and Sentence selection. In doing so, we hope to standardize our scores in order to not get bogged down by redundancies. We will draw upon Vasu et all's paper on BioAMA[5] and possibly collaborate with them to find possibilities to enhance their pipeline.

In particular, we test on the ideal answer and exact answer separately. The ideal answer type will evaluated by the question level ROUGE scores using the different combinations ranking functions at the ranker level as well as different methodology of the sentence selection procedure. As for the exact answer which is described in the below section. Finally, we use statistics significant test to analysis the error.

We designed a new accuracy metric based on hard and soft measure to define the number of exact answers which were found in the predicted answer. This analysis was done for yes-no, list and factoid type questions.

\section{Analysis for Summary Type Question}
In this part, we test our pipeline for the summary type question. We will first show the results on both different sentence selection modules and feature extractors, and then perform error analysis.

First, we show our results on different sentence selection modules; the greedy selection serves as the baseline, which greedily selects top $k$ sentences according to their similarity score between the question. We compare the baseline with MMR Module. Note that in this part we fixed the feature selection module to the Jaccard similarity feature. The results are shown in Table~\ref{tbl:sentence}. We could clearly see that MMR Module is better than the greedy baseline, which verify our hypothesis that reducing the redundancy is important for the summary questions.

\begin{table}[t!]
\centering
\caption{Greedy Selection vs MMR.}
\label{tbl:sentence}
\begin{tabular}{|l|l|l|l|l}
\cline{1-4}
\textbf{k}  & \textbf{Precision} & \textbf{Recall} & \textbf{F1}    &  \\ \cline{1-4}
1  & 0.784     & 0.222  & 0.292 &  \\ 
2  & 0.623     & 0.373  & 0.38  &  \\ 
3  & 0.522     & 0.485  & 0.398 &  \\ 
4  & 0.469     & 0.553  & 0.407 &  \\ 
5  & 0.415     & 0.612  & 0.415 &  \\ 
6  & 0.388     & 0.658  & 0.406 &  \\ 
7  & 0.376     & 0.671  & 0.409 &  \\ 
8  & 0.366     & 0.662  & 0.396 &  \\ 
9  & 0.358     & 0.674  & 0.392 &  \\ 
10 & 0.362     & 0.657  & 0.385 &  \\ \cline{1-4}
\end{tabular}
\end{table}

Secondly, we show the results on different feature extractor modules: Jaccard similarity, Dice similarity, Tf-Idf similarity and the combined feature. The results are shown in Table~\ref{tbl:feature}. It could be observed that overall the Jaccard feature has the best Rouge score. Our experiment with the pre-trained Dependency Parser resulted in increase in precision of the Rouge Score, however, it did not result in improvement in Rouge-F1 score. We attribute this result to cosine and td-idf bringing semantic similarity into picture which is more effective than syntactical similarity as offered by the parser, as shown in Table~\ref{tbl:parser}.

\begin{table}[]
\centering
\caption{Results on Dependency Parser.}
\begin{tabular}{|l|l|l|l|}
\hline
\textbf{Method}        & \textbf{Precision} & \textbf{Recall} & \textbf{F1} \\ \hline
Original   & 0.415     & 0.612  & 0.415  \\ 
Parser     & 0.422     & 0.591  & 0.405  \\ \hline
\end{tabular}
\label{tbl:parser}
\end{table}

\begin{table}[]
\centering
\caption{Comparison of Feature Extractors.}
\label{tbl:feature}
 \resizebox{0.98\columnwidth}{!}{
\begin{tabular}{|l|l|l|l|l|}
\hline
\textbf{k}  & \textbf{Jaccard} & \textbf{Dice}  & \textbf{Tf-Idf} & \textbf{Combined} \\ \hline
1  & 0.292   & 0.321 & 0.327  & 0.307    \\ 
2  & 0.38    & 0.379 & 0.386  & 0.374    \\ 
3  & 0.398   & 0.4   & 0.403  & 0.391    \\ 
4  & 0.407   & 0.412 & 0.396  & 0.412    \\ 
5  & 0.415   & 0.397 & 0.396  & 0.409    \\ 
6  & 0.406   & 0.4   & 0.394  & 0.404    \\ 
7  & 0.409   & 0.398 & 0.388  & 0.399    \\ 
8  & 0.396   & 0.394 & 0.393  & 0.398    \\ 
9  & 0.392   & 0.395 & 0.389  & 0.398    \\ 
10 & 0.385   & 0.393 & 0.388  & 0.402    \\ \hline
\end{tabular}}
\end{table}

Now we take a deeper look in our experiments. The first question raised is how we choose the number of candidate sentences. In order to choose the best $k$, we resort to the Wilcoxon significance test to determine under which number our experiment has statistics significance. We first test on the sentence selection experiment and show the results in Table~\ref{tbl:error}. Obviously, we can see that we have experiment significant when $k$ is range from $4$ to $7$.

After we decide $k$, we could also use significant test to justify our results on the feature extractor experiments. To be more specific, we compare the best feature: Jaccard with the others in Table~\ref{tbl:dice}, Table~\ref{tbl:tf}, and Table~\ref{tbl:combine}.
\begin{table}[]
\centering
\caption{Wilcoxon Significant Test on Sentence Selection.}
\label{tbl:error}
 \resizebox{0.98\columnwidth}{!}{
\begin{tabular}{|l|l|l|l|l|}
\hline
\textbf{k}& \textbf{Greedy} & \textbf{MMR}   & \textbf{Mean-diff} & \textbf{Wilcoxon}\\ \hline
2         & 0.36   & 0.374 & 0.014     & 0.379       \\ 
3         & 0.459  & 0.485 & 0.026     & 0.069       \\ 
4         & 0.528  & 0.553 & 0.025     & 0.17        \\ 
5         & 0.573  & 0.612 & 0.039     & 0.018       \\ 
6         & 0.623  & 0.658 & 0.035     & 0.02        \\ 
7         & 0.635  & 0.671 & 0.036     & 0.005       \\ 
8         & 0.653  & 0.662 & 0.009     & 0.652       \\ 
9         & 0.651  & 0.674 & 0.023     & 0.039       \\ 
10        & 0.654  & 0.6   & -0.054    & 0.517       \\ \hline
\end{tabular}}
\end{table}

\begin{table}[]
\centering
\caption{Jaccard vs Dice.}
\label{tbl:dice}
 \resizebox{0.98\columnwidth}{!}{
\begin{tabular}{|l|l|l|l|l|}
\hline
\textbf{k} & \textbf{Dice}  & \textbf{Jaccard} & \textbf{Mean-diff} & \textbf{Wilcoxon} \\ \hline
4 & 0.513 & 0.553   & 0.04      & 0.004    \\ 
5 & 0.571 & 0.612   & 0.041     & 0.003    \\ 
6 & 0.614 & 0.658   & 0.044     & 0.001    \\ 
7 & 0.635 & 0.671   & 0.036     & 0.002    \\ \hline
\end{tabular}}
\end{table}

\begin{table}[]
\centering
\caption{Jaccard vs Tf.}
\label{tbl:tf}
\begin{tabular}{|l|l|l|l|l|}
\hline
\textbf{k} & \textbf{Tf}    & \textbf{Jaccard} & \textbf{Mean-diff} & \textbf{Wilcoxon} \\ \hline
4         & 0.55  & 0.553   & 0.003     & 0.982    \\ 
5         & 0.594 & 0.612   & 0.018     & 0.192    \\ 
6         & 0.63  & 0.658   & 0.028     & 0.043    \\ 
7         & 0.648 & 0.671   & 0.023     & 0.036    \\ \hline
\end{tabular}
\end{table}

\begin{table}[]
\centering
\caption{Jaccard vs Combined.}
\label{tbl:combine}
 \resizebox{0.98\columnwidth}{!}{
\begin{tabular}{|l|l|l|l|l|}
\hline
\textbf{k} & \textbf{Combined} & \textbf{Jaccard} & \textbf{Mean-diff} & \textbf{Wilcoxon} \\ \hline
4 & 0.522    & 0.553   & 0.031     & 0.01     \\
5 & 0.593    & 0.612   & 0.019     & 0.136    \\ 
6 & 0.635    & 0.658   & 0.023     & 0.121    \\
7 & 0.652    & 0.671   & 0.019     & 0.15     \\ \hline
\end{tabular}}
\end{table}

We could see that the Jaccard indeed is supported by the significant test for its better performance against others. However, we note that it becomes more unstable for the combine feature, we think it is due to the high variance of the experiment results. 

\section{Analysis of Exact answer based evaluation metric}

The table \ref{tbl:yesno} shows results for the yesno type questions across different similarity metrics such as cosine, dice, jaccard and all combined for k value ranging from 4-7. We notice that all of the metrics indicate that yesno questions dont find yes or no in the predicted answer. Jaccard similarity measure on the other hand does a better job of predicting the yes or no in the predicted answer on 2 out of the 21 questions which were analyzed in the validation set. 

\begin{table}[]
\centering
\caption{Results for Yes/No type questions}
\label{tbl:yesno}
 \resizebox{0.98\columnwidth}{!}{
\begin{tabular}{|l|l|l|l|l|l|l|l|l|}
\hline
\textbf{k} & \textbf{\thead{Cosine\\  soft\\  measure }}& \textbf{\thead{Cosine\\  hard \\ measure }} & \textbf{\thead{Dice\\  soft\\  measure}} & \textbf{\thead{Dice\\ hard\\ measure}} & \textbf{\thead{Jaccard\\ Soft\\ measure}} & \textbf{\thead{Jaccard\\ hard\\ measure}} & \textbf{\thead{All\\ Soft\\ measure }} & \textbf{\thead{All\\ Hard\\ measure}} \\ \hline
4 & 0.021    & 0.021   & 0.021     & 0.021 & 0 & 0 & 0.021 & 0.021 \\
5 & 0.021   & 0.021   & 0.021     & 0.021  & 0.04 & 0.04 & 0.021 & 0.021\\ 
6 & 0.021    & 0.021  & 0.021     & 0.021  & 0.042  & 0.042 & 0.021 & 0.021\\
7 & 0.021    & 0.021   & 0.021     & 0.021  & 0.042  & 0.042 & 0.021 & 0.021\\ \hline
\end{tabular}}
\end{table}

The table \ref{tbl:list} shows results for the list type questions across different similarity metrics and for k value ranging from 4-7. We observe a huge gap in the hard measure and soft measure metrics scores. This can be due to the fact that the list type question has many options of the exact answer present as compared to other type questions. Cosine measure performs well for k = 4 and 6 whereas Jaccard soft measure performs well for k = 5 and 7. The ratio of the number of questions getting atleast one of the output correct is greater than 50 percent most of all the cases which is a very good indicator of correct output for such question types.   

\begin{table}[]
\centering
\caption{Results for List type questions}
\label{tbl:list}
 \resizebox{0.98\columnwidth}{!}{
\begin{tabular}{|l|l|l|l|l|l|l|l|l|}
\hline
\textbf{k} & \textbf{\thead{Cosine\\ soft\\ measure}}& \textbf{\thead{Cosine\\  hard\\ measure}} & \textbf{\thead{Dice\\ soft\\ measure}} & \textbf{\thead{Dice\\ hard\\ measure}} & \textbf{\thead{Jaccard\\ Soft\\ measure}} & \textbf{\thead{Jaccard \\hard\\ measure}} & \textbf{\thead{All\\ Soft\\ measure}} & \textbf{\thead{All\\ Hard\\ measure }} \\ \hline
4 & 0.61    & 0.1   & 0.51     & 0.085  & 0.57 & 0.085 & 0.55 & 0.085 \\
5 & 0.59   & 0.106   & 0.531     & 0.106  & 0.61 & 0.106 & 0.48 & 0.045\\ 
6 & 0.578    & 0.085  & 0.55     & 0.085  & 0.55 & 0.106 & 0.53 & 0.063\\
7 & 0.591    & 0.106   & 0.55     & 0.085  & 0.61  & 0.107 & 0.55 & 0.085\\ \hline
\end{tabular}}
\end{table}


The table \ref{tbl:factoid} shows results for the factoid type questions across different similarity metrics such as cosine, dice, jaccard and all combined for k value ranging from 4-7. The factoid type questions has a value of the soft and hard measure very close to each other for a given type of similarity metric. This could be due to the fact that exact answer has only one exact answer in most of these type of questions. Further, cosine measure and all measure perform well on 3 of the 4 different k values we tested indicating that the factoid type questions can we answered reasonably well with exact type present in the predicted answers. 

\begin{table}[]
\centering
\caption{Results for Factoid type questions}
\label{tbl:factoid}
 \resizebox{0.98\columnwidth}{!}{
\begin{tabular}{|l|l|l|l|l|l|l|l|l|}
\hline
\textbf{k} & \textbf{\thead{Cosine  \\ soft \\measure}}& \textbf{\thead{Cosine \\ hard \\measure}} & \textbf{\thead{Dice \\soft \\measure}} & \textbf{\thead{Dice \\hard \\measure}} & \textbf{\thead{Jaccard \\Soft\\ measure}} & \textbf{\thead{Jaccard\\ hard \\measure}} & \textbf{\thead{All\\Soft\\measure}} & \textbf{\thead{All\\Hard\\measure}} \\ \hline
4 & 0.66    & 0.52   & 0.61     & 0.47  & 0.61 & 0.5 & 0.64 & 0.47 \\
5 & 0.59   & 0.45   & 0.61     & 0.136  & 0.54 & 0.42 & 0.64 & 0.54\\ 
6 & 0.61    & 0.5  & 0.61     & 0.121  & 0.61 & 0.47 & 0.61 & 0.52\\
7 & 0.61    & 0.45   & 0.61     & 0.15  & 0.57  & 0.452 & 0.61 & 0.5\\ \hline
\end{tabular}}
\end{table}


\section{Results analysis - Sample output}

\subsection{Sample - 1}
\textbf{Question}- What colonoscopy findings have been reported in autism?

\textbf{Ground Truth Answer} - Endoscopy trials have demonstrated a higher prevalence of nonspecific colitis, lymphoid hyperplasia and focally enhanced gastritis compared with controls.

\textbf{Predicted answer (Greedy Approach)} - Intestinal mucosal pathology, characterized by ileo-colonic lymphoid … variant of inflammatory bowel and stomach, has been reported in children with developmental disorders. Autistic enterocolitis : fact or fiction?

\textbf{Predicted answer (MMR)} - Developmental diagnoses were autism ( 50 patients ) …  and bowel symptoms endoscopy \textbf {trials have demonstrated a higher prevalence of nonspecific colitis, lymphoid hyperplasia and focally enhanced gastritis compared with controls.}

In this particular example, we compare the summary generated from the Greedy sentence selection module and the MMR module: We note that the summary generated by the MMR algorithm is more linguistically reasonable, while the one from greedy approach is like a random combination of sentences. This interesting phenomenon supports the hypothesis that MMR can effectively reduce the redundancy of the candidates, which is beneficial for the tiler module afterward. Moreover,the Rouge score (recall) for them are $0.84$ and $0.97$, respectively. The reason is that MMR could select more diverse candidates, which actually enhances the probability for including the golden answer. On the other hand, the sentences select by Greedy approach is similar to each other, which predicts highly unstable results.

\subsection{Sample - 2}
\textbf{Question} - GV1001 vaccine targets which enzyme? \\
\textbf{Predicted answer} - Background : we aimed to assess the efficacy and safety of sequential or simultaneous telomerase ( htert ) derived peptide, gv1001, has been developed as a vaccine against various cancers.A study was conducted to investigate safety and immunogenicity in non-resectable pancreatic carcinoma patients using a 16-amino acid telomerase peptide ( GV1001 ) for vaccination in combination with GM-CSF and gemcitabine as first line treatment.Gemcitabine and capecitabine with or without telomerase peptide vaccine GV1001 in patients with locally advanced or metastatic pancreatic cancer ( TeloVac ) : an open-label, randomised, phase 3 trial.Peptide GV1001 is a peptide vaccine representing a 16-amino acid \textbf{human telomerase reverse transcriptase sequence}, which has been reported to possess potential antineoplastic and anti-inflammatory activity.A reverse-transcriptase-subunit of telomerase vaccination ( gv1001 ) in combination with chemotherapy appeared to be safe but the immune responses were weak and transient.GV1001 is a 16-amino-acid vaccine peptide derived from the human telomerase reverse transcriptase sequence. \\
\textbf{Exact answer} - human telomerase reverse transcriptase \\

We observe that the answer generated in our case is consisting of on an average 200 words which is large as compared with the ideal answer word count. This provides us insights into having a better tiling module as the tiler currently is concatenating the outputs or just using a LSTM network~\cite{hochreiter1997long} output to generate the final output. Having constraints on the number of words one can expect for this question would help improve the quality of the predicted answer.  

\subsection{Sample - 3}
\textbf{Question} Is mitofusin 2 a receptor for parkin?\\
\textbf{Ideal answer} Yes, Mfn2 functions as a mitochondrial receptor for Parkin.\\ 
\textbf{Predicted answer} Mfn2 functions as a mitochondrial fusion protein mitofusin 2 serves as a receptor for parkin translocation to damaged mitochondria.Mfn2 functions as a mitochondrial receptor for Parkin and is required for quality control of cardiac mitochondria.Recent work demonstrates that a phosphorylated form of the mitochondrial fusion protein mitofusin 2 serves as a receptor for parkin recruitment to damaged mitochondria.Mitofusin 1 and mitofusin 2 are ubiquitinated in a PINK1 manner upon induction of mitophagy we show that the mitochondrial outer membrane guanosine triphosphatase mitofusin ( Mfn ) 2 mediates Parkin recruitment to damaged mitochondria.\\
\textbf{Exact answer} Yes\\
For this question, we notice a useful trend which is predicted in the predicted answer from our pipeline. We observe that the predicted answer contains the words Mitofusin 2 serves as a receptor for mitochondrial receptor. This means that the pipeline is understanding the question which is being asked and infact this predicted answer is close to ideal answer. Despite the question of yesno answer type not being able to predict exact answer of yes or no, one can easily infer that the answer being predicted correctly in our pipeline. Thus, work is needed here to understand the meaning of the predicted answer and also append yes or no to the predicted answer as a post processing step to predict yes or no. Such examples indicate that our pipeline is indeed working well for the summary type questions. We notice here as well the length of the ideal answer is very small as compared to our answer which prints out a summary producing content which is logical but not concise. 

\subsection{Sample - 4}
\textbf{Question}  Which are the DNA (cytosine-5-)-methyltransferases inhibitors?\\
\textbf{Ideal answer} DNA (Cytosine-5-)-methyltransferases are a family of enzymes that methylate DNA at the C5 position of cytosine residues. Given that methylation of tumour suppressor gene promoters leads to carcinogenesis, inhibition of DNA (Cytosine-5-)-methyltransferases is a promising strategy for the treatment of cancer. There are several inhibitors of DNA (Cytosine-5-)-methyltransferases that uses different modes of action: 5-azacytidine (5-aza-CR, Vidaza®), 5-azadeoxycytidine (5-aza-CdR, decitabine, Dacogen®), 5-azacytosine (ZCyt), 5-fluorodeoxycytidine (FdC), 5,6-dihydro-5-azacytosine (DZCyt), 4'-thio-2'-deoxycytidine, hydralazine, 2-(1H)-pyrimidinone riboside (zebularine), 2-(1H)-pyrimidinone (zebularine aglycon), procaine, procainamide, psammaplin A, and RSC133, a new synthetic derivative of indoleacrylic acid/indolepropionic acid." \\
\textbf{Predicted answer} A number of dna ( cytosine-5- ) -methyltransferases by \textbf{5-azacytosine} is likely to involve methyl transfer to the inhibitor.Although \textbf{5-Aza-CdR} is known to be incorporated into DNA and inhibit DNA ( cytosine-5 ) -methyltransferases, the precise mechanisms underlying the drug's antineoplastic activity remain unclear.The inhibitors showed selectivity for c5 dna ( cytosine-c5 ) methyltransferase containing 5-azacytosine residues at specific sites.Although \textbf{5-aza-cdr} is known to be incorporated into dna ( cytosine-c5 ) methyltransferase containing 5-azacytosine residues at specific sites.Cooperative action of hdaci and dna methylation inhibitors ( dnmti ) has been reported, making combined treatment an attractive choice for cancer therapy.The mechanism of inhibition of dna methyltransferase inhibitors are known to reactivate silenced genes; including \textbf{5-azacytidine} and \textbf{2- ( 1h ) -pyrimidinone riboside} ( \textbf{zebularine} ).Synthesis of oligonucleotide inhibitors of dna ( cytosine-c5 ) methyltransferase containing \textbf{5-azacytosine} is likely to involve methyl transfer to the inhibitor.\\ 
\textbf{Exact Answer}'5-azacytidine', '5-aza-CR', 'Vidaza®', '5-azadeoxycytidine', '5-aza-CdR', 'Dacogen®', "5-aza-2'-deoxycytidine", 'Decitabine', '5-azacytosine', 'ZCyt', '5-fluorodeoxycytidine', 'FdC', '5,6-dihydro-5-azacytosine', 'DZCyt', "4'-thio-2'-deoxycytidine", 'hydralazine', '2-(1H)-pyrimidinone riboside', 'zebularine', '2-(1H)-pyrimidinone', 'zebularine aglycon', 'procaine', 'procainamide', 'RSC133', 'Psammaplin A'\\ 

For the above difficult question, we notice that there lots of exact answer samples captured in the predicted answer which implies our model pipeline is performing well. Further the sentence meaning of the predicted answer is close to the ideal answer. We have highlighted the exact answer which are found in our predicted answer for reader to evaluate the prediction received. Ideal answer is predicting a list of DNA inhibitors in this case and one can account for that fact in the text and search for such key phrases on what to look for can be critical for developing a better list type question answering model. 

\section{Lessons Learned}
We had several positive results from this project. First outcome was that we understood the BioASQ challenge task and built an end to end pipeline for predicting the ideal answer for it. We utilized the software engineering concepts which were taught in class to present our results at various stages: pipeline design, error analysis, and code refine. Moreover, by using statistics significant test to analyze our results, We discover that Wilcoxon test is the most suited significance test for this particular task. Our comparative analysis also provided insights into the different types of questions as well - for instance, we conclude that this pipeline is not suitable for yes/no question answering and more work needs to be done to provide a predicted answer which is more closer to the ideal answer. Most importantly, We worked on developing a modular designed code which was also taught concept during class and will be beneficial in our future career, too.

\section{Future Work/Extensions}
There are a few directions we think are important to pursue in future. First is to improve the ranker module with additional ranker pipelines. Fusion based approaches could be useful to apply on this task and see the results which can be provided by the late fusion approach on not only the final sentence tiling output but to the output of the ranker module as well. Moreover, including word representation like word2vec~\cite{mikolov2013distributed} or Glove vector~\cite{pennington2014glove} as the additional feature is also a direction that we will try in the future. Furthermore, yes no questions needs to be addressed differently using methods where one can convert the question into a assertion statement and view the output differences between the ideal answer and predicted answer.
One possible solution is to train a binary classifier, whose inputs are the question along with the predicted summary, while the output is Yes/No. With the development of recurrent neural network~\cite{wu2016google,arora2016simple}, we are looking forward to exploring this pipeline in the future.  

\section{Summary}
We designed an end-to-end pipeline improving upon a reference implementation of the BioASQ pipeline. Our experiments were focussed towards improving the feature extractor part of the pipeline and improving the understanding of overall pipeline. We obtained a significant result conclusion such as range of k value which is statistically significant. Our results provided insights into using wilcoxon test for significance testing in this task. Further, we developed and analyzed exact answer based accuracy metric to evaluate performance of our pipeline. We recommend future directions for using our work to develop better systems on BioASQ challenge tasks and draw conclusions from the question answering task. 

\bibliographystyle{acl_natbib}
\bibliography{acl2017}
\end{document}